%% file: main.tex
\documentclass[letterpaper]{article} 
\usepackage{aaai2026}  
\usepackage{times}  
\usepackage{helvet}  
\usepackage{courier}  
\usepackage[hyphens]{url}  
\usepackage{graphicx} 
\usepackage{natbib}  
\usepackage{caption} 
\usepackage{algorithm}
\usepackage{algorithmic}
\usepackage{newfloat}
\usepackage{listings}
\DeclareCaptionStyle{ruled}{labelfont=normalfont,labelsep=colon,strut=off} 
\floatstyle{ruled}
\newfloat{listing}{tb}{lst}{}
\floatname{listing}{Listing}

\usepackage{amsmath}
\usepackage{booktabs} 
\usepackage{framed}
\usepackage{xurl}
\usepackage{color}
\usepackage{multirow}

\title{How English Print Media Frames Human–Elephant Conflict in India}
\author {
   Anonymous submission
}
\author {
    Bonala Sai Punith\equalcontrib\textsuperscript{\rm 1},
    Salveru Jayati\equalcontrib\textsuperscript{\rm 1},
    Garima Shakya\textsuperscript{\rm 1},
    Shubham Kumar Nigam\textsuperscript{\rm 2},
}
\affiliations {
    \textsuperscript{\rm 1}Indian Institute of Technology Palakkad, India\\
    \textsuperscript{\rm 2}University of Birmingham Dubai, United Arab Emirates \\
   bonalasaipunith@gmail.com, salverujayati@gmail.com, garima@iitpkd.ac.in,
    shubhamkumarnigam@gmail.com
}

\usepackage{bibentry}

\begin{document}

\maketitle

\input{abstract}


\input{intro}

\input{related_work}

\input{task_description}

\input{dataset}

\input{methods}

\input{experimental_setup}

\input{results_and_analysis}

\input{discussion_combined}

\input{conclusion_and_future_work}

\bibliography{references, software}

\appendix
\input{appendix}
\end{document}

%% file: abstract.tex
\begin{abstract}
Human-elephant conflict (HEC) is rising across India as habitat loss and expanding human settlements force elephants into closer contact with people. While the ecological drivers of conflict are well-studied, how the news media portrays them remains largely unexplored. This work presents the first large-scale computational analysis of media framing of HEC in India, examining 1,968 full-length news articles consisting of 28,986 sentences, from a major English-language outlet published between January 2022 and September 2025. Using a multi-model sentiment framework that combines long-context transformers, large language models, and a domain-specific Negative Elephant Portrayal Lexicon, we quantify sentiment, extract rationale sentences, and identify linguistic patterns that contribute to negative portrayals of elephants. Our findings reveal a dominance of fear-inducing and aggression-related language. Since the media framing can shape public attitudes toward wildlife and conservation policy, such narratives risk reinforcing public hostility and undermining coexistence efforts. By providing a transparent, scalable methodology and releasing all resources through an anonymized repository, this study highlights how Web-scale text analysis can support responsible wildlife reporting and promote socially beneficial media practices.
\end{abstract}


%% file: intro.tex
\section{Introduction}

Human--elephant conflict (HEC) has emerged as one of the most persistent conservation and socio-political challenges across South Asia. India hosts the world’s largest population of Asian elephants, with an estimated 25,000--30,000 individuals distributed across roughly 163,000 km$^2$ of heterogeneous landscapes~\cite{roy2025long}. Between 2019 and 2024 alone, the country recorded 2,829 human fatalities linked to elephant encounters, while 528 elephants died due to electrocution, train collisions, poisoning, and poaching~\cite{GoI_Lok_sabha2024}. Although ecological and socio-economic drivers of HEC have been widely studied~\cite{roy2025long}, the analysis of print media portrayal of the elephants remains comparatively underexplored.

For many urban and semi-urban readers, news reporting is the primary gateway to understanding wildlife-related events. Media narratives thus influence attitudes~\cite{shanahan2011narrative}, shape public sentiment, and may inadvertently fuel fear, antagonism, or misinformation \cite{mccombs2005look, muter2013australian}. Sensational framing, often using emotionally charged or anthropomorphic labels such as \textit{“menace”, “rogue elephant”, “killer jumbo”, “attack”,} or \textit{“rampage”}, can distort the ecological realities of conflict and reinforce skewed perceptions of elephants as aggressive actors rather than animals responding to habitat degradation or blocked migratory corridors~\cite{de2012human,MAURER2024110391}. Such narratives can undermine coexistence-oriented conservation strategies and influence policy discourse.

\begin{figure}[t]
    \centering
    \includegraphics[width=0.90\linewidth]{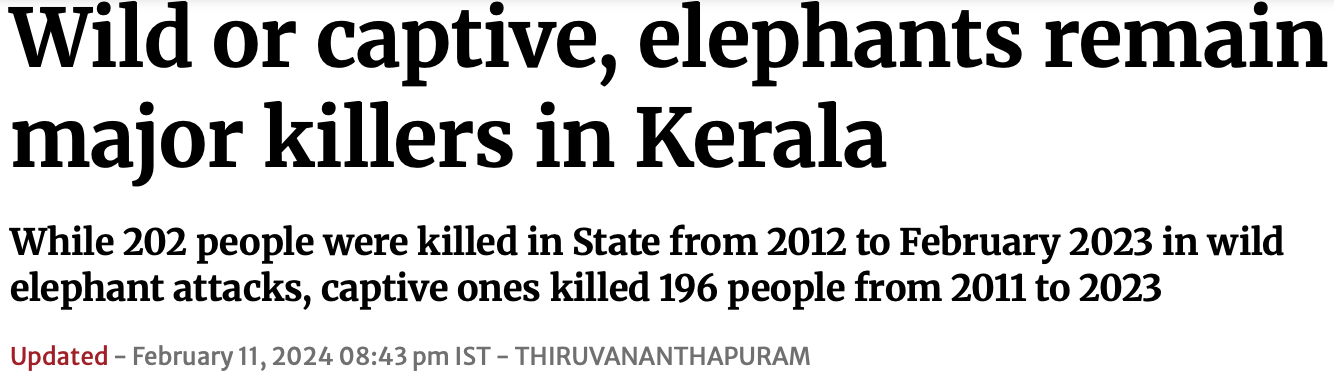}
    \\\vspace{10pt}
\includegraphics[width=0.90\linewidth]{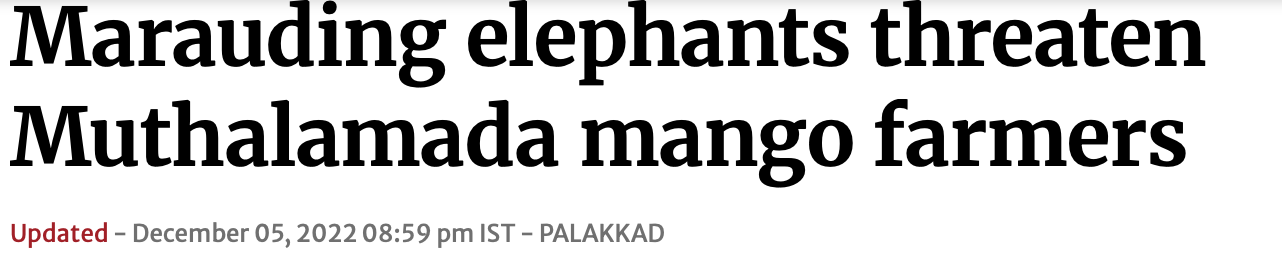}
    \caption{
These narratives attribute intentionality to animals, transforming ecological events into stories of attack and defense. This framing positions humans as orderly and rational while depicting elephants as irrational and violent. Such framing positions humans as orderly and rational, while painting elephants as irrational and violent \cite{cheerotha2025}.
    }
    \label{fig:media_lens}
\end{figure}

In contrast, balanced reporting that situates conflict within broader ecological and socio-political contexts can foster empathy, promote coexistence, and support mitigation interventions. Recognizing this, the Ministry of Environment, Forest and Climate Change has periodically advised the media to avoid sensational or dehumanizing language when reporting on elephants~\cite{IE_2021}.  Yet anecdotal evidence suggests that such advisories have had limited practical effect. 
\begin{figure}
    \centering
    
\includegraphics[width=0.8\linewidth]{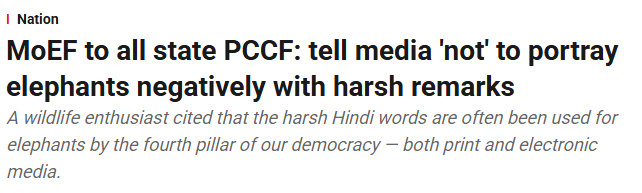}
    \caption{A news article stating the advisory by the Ministry of Environment and Forest (MoEF) in September 2021~\cite{IE_2021}. }
    
\includegraphics[width=0.8\linewidth]{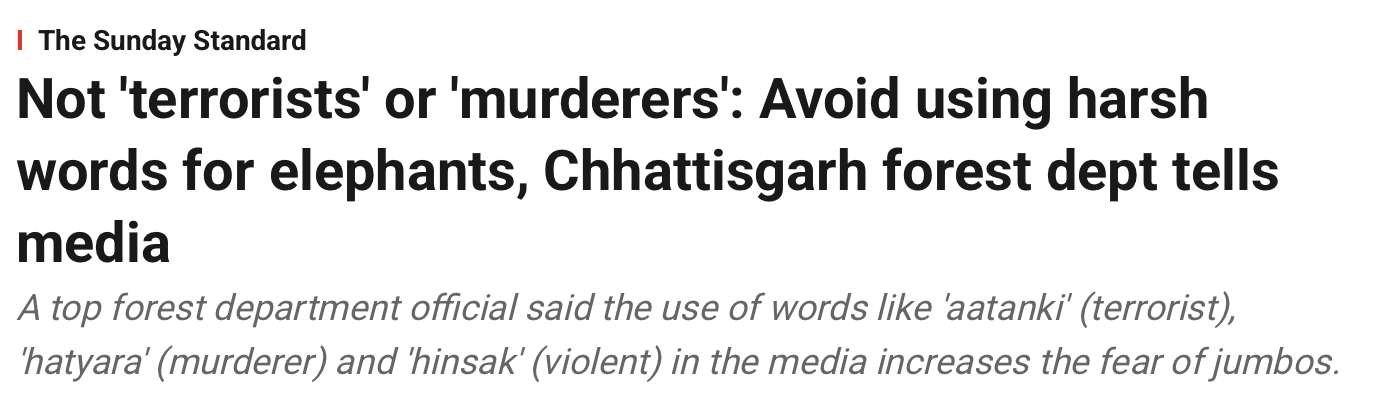}
    \caption{A news article stating the advisory by Chhattisgarh Forest Department to the media in March 2025~\cite{IE_2025}.}

\label{fig:placeholder}
\end{figure}


The motivation of this study is to systematically analyze how media frames elephants in news reporting and whether sentiment patterns exhibit negative bias at scale. We propose a multi-model evaluation pipeline that integrates transformer models, long-context architectures, rule-based lexicons, and instruction-tuned large language models (LLMs) to assess sentiment and extract rationale sentences from full-length articles. Beyond polarity classification, our framework offers a scalable methodology for assessing wildlife portrayals in high-stakes conservation contexts. This aligns well with the objectives of AIES by demonstrating how computational tools can support social good, improve media accountability, and contribute to evidence-based approaches for fostering human–wildlife coexistence.

\subsection*{Challenges}
Analyzing sentiment and narrative framing in long-form news articles presents several methodological challenges. 
\begin{itemize}
    \item First, sentiment toward elephants is frequently implicit rather than overt; articles may maintain a neutral journalistic tone while framing elephants as threats through selective verb choice or event emphasis. This implicit sentiment is difficult for traditional sentiment models, which are tuned primarily for explicit affect.
    \item Second, acquiring a comprehensive dataset on HEC reporting is non-trivial in itself. Although organizations like MediaCloud\footnote{\url{https://search.mediacloud.org/} An open-source platform
for media analysis.} provided article URLs for selected outlets, additional scraping, filtering, and cleaning were required to identify HEC-relevant articles and remove noise.
\item Third, the absence of large-scale human-annotated ground truth complicates evaluation. Standard supervised metrics (accuracy, F1) are not directly applicable. Thus, we rely on alternative measures such as cross-model agreement and expert justification assessment, which provide meaningful but indirect evaluation signals.
\end{itemize}

While popular reporting tends to foreground dramatic incidents such as crop destruction, attacks on villagers, or ``rogue elephant'' events, such depictions overlook the ecological and socio-political factors that precipitate conflict. This mismatch between narrative emphasis and ecological reality motivates our linguistic and sentiment-based investigation of news articles. By systematically quantifying framing patterns, we aim to uncover the mechanisms through which media representations shape public understanding of HEC.

\subsection*{Contributions}
This paper makes the following key contributions:
\begin{itemize}
    \item \textbf{First quantitative analysis of Indian media framing of HEC}, covering 1,968 articles from one of the most-read English-language national outlets.
    \item \textbf{A multi-model sentiment analysis framework} integrating LLMs, long-context transformers, rule-based systems, and lexicon-driven detectors to capture both explicit and implicit sentiment cues (Table~\ref{tab:hyperparameter} and \ref{tab:prompt}).
    \item \textbf{A novel Negative Elephant Portrayal Lexicon (NEPL)} combining lexical, metaphorical, and fear-inducing expressions grounded in expert review (Table~\ref{tab:nepl}).
    \item \textbf{Comprehensive bias assessment} across the timeline, fatal vs. non-fatal incidents, assessing and comparing AI models with the expert annotations, revealing systemic negative framing trends. (Figures~\ref{fig:pie_chart},~\ref{fig:model_comp},~\ref{fig:model_agreement_comp}, Table~\ref{tab:models_vs_annotator_perf}).
    \item \textbf{Sentence-level rationale extraction} using LLMs to identify narrative drivers behind sentiment decisions (Table~\ref{tab:llm_vs_annotator_perf}, ~\ref{tab:llm_reasoning_agreement}).
\end{itemize}

%% file: related_work.tex
\section{Related Work}
\label{sec:related_work}

Human–elephant conflict and its representation in public discourse have been examined from ecological, sociological, and, more recently, computational perspectives. The literature demonstrates that media narratives significantly shape public perceptions of wildlife, influence conservation attitudes, and affect policy decisions \cite{mccombs2005look, muter2013australian}. Conservation media studies show that news framing can amplify fear, sensationalize conflict, or alternatively, promote coexistence and empathy. For Asian elephants, this relationship between media portrayal and public perception is especially pronounced, given the species’ dual cultural status as both revered and feared.

\textbf{Analysis of Media Coverage.} A comprehensive study by \cite{maurer2024polarized}, utilizing topic modeling, examines how Asian elephants are framed in 11,000 global news articles published over 13 years through 2019. The findings indicate that episodic, event-driven coverage—typically reporting crop damage, injuries, or fatalities—relies heavily on emotional and spatially grounded narratives, which heighten public perceptions of danger. In contrast, thematic articles employ a more institutional lexicon focused on policy, conservation, and long-term ecological trends. This dichotomy demonstrates that different reporting styles can produce divergent public perceptions of human–elephant conflict (HEC) and suggests that media narratives often emphasize individual, dramatic events rather than structural drivers of conflict. Earlier studies by Barua \cite{barua2010whose} and Doyle et al. \cite{de2012human} similarly used content analysis to show that media disproportionately emphasize negative encounters, contributing to a narrative of elephants as unpredictable or dangerous.

Media-driven fear amplification extends beyond elephants. Research on large carnivores, spiders, and sharks demonstrates similar patterns, where fear-based framings distort public risk perception \cite{nanni2020social, mammola2022global, okpala2024framework}.
Parallel research has begun to explore the influence of social media and user-generated content on perceptions of wildlife. Analyses of YouTube videos featuring elephants \cite{ladue2025representations} and other charismatic species demonstrate that elephant stress behavior correlates with protected-area status and human activity. These findings underscore a feedback loop between sensational content and user engagement, raising concerns about the impact of digitally mediated narratives on conservation practices and tourist expectations.

\textbf{Ecological Research on HEC.} In addition to media studies, ecological research on human–elephant conflict has addressed long-term spatial and temporal dynamics. Recent investigations have examined mortality trends \cite{ekuban2025time}, conflict hotspots \cite{roy2025long}, and spatial drivers such as habitat fragmentation and agricultural expansion \cite{goswami2015mechanistic, bal2011elephants} across various regions of India. Studies focused on Asian elephants \cite{menon2019population, prakash2019reasons} have evaluated mitigation strategies, including early warning systems, community engagement, and the development of elephant corridors. Reviews from Indonesia \cite{kuswanda2022can} emphasize the importance of coexistence-based approaches rooted in ecological restoration, community empowerment, and corridor development. Additional research on Asian and African elephants \cite{mumby2018taking} highlights the necessity of understanding elephants’ decision-making processes to design humane and effective interventions.

\textbf{The Role of Large Language Models in Conservation.} Recently, large language models (LLMs) have been utilized as tools for analyzing wildlife-related issues, including trafficking \cite{10.1145/3725256}.
\citet{santangeli2024large} demonstrated that LLMs can debunk false and sensational wildlife news. \citet{rathi2025navigating} provides a comprehensive report indicating that technologies such as machine learning, computer vision, and natural language processing improve species identification, population estimation, and the monitoring of poaching activities.

Despite extensive ecological and sociological research, a significant gap remains at the intersection of conservation science, media studies, and computational linguistics. Very few works employ modern natural language processing (NLP) and LLM-based techniques to systematically examine how news media constructs narratives about elephants, particularly regarding how sentiment, lexical choices, and negative portrayals contribute to broader societal fear or misperceptions. This study addresses this gap by providing a large-scale, multi-model sentiment and rationale analysis of Indian news articles concerning human–elephant conflict, offering new insights into the linguistic mechanisms that can shape public perceptions of elephants.

%% file: task_description.tex
\section{Problem Description}
\label{sec:task_description}

The core goal of this study is to analyze how news articles portray elephants in the context of HEC by examining the overall sentiment expressed in the text and identifying the specific sentences that contribute to that sentiment. Rather than decomposing articles into structural components (such as headline or body), we treat each article as a single, holistic document and perform sentiment estimation directly on the full text. Formally define the problem given a full news article $a_i$, determine its sentiment toward elephants, and extract the sentences that most likely contribute to that sentiment.


Let the dataset consist of $N$ full-length news articles:
\[
\mathcal{D} = \{ a_1, a_2, \dots, a_N \}, \qquad a_i \in \mathcal{T},
\]
where $\mathcal{T}$ is the space of raw article texts.  
For each article $a_i$, the task is to determine:

\begin{enumerate}
    \item \textbf{Article Sentiment:}  
    A polarity score indicating whether the article expresses an overall
    \textit{positive}, \textit{neutral}, or \textit{negative} sentiment toward elephants.
    
    \item \textbf{Rationale Sentences:}  
    A set of sentences within the article that likely contribute to or justify the predicted sentiment.
\end{enumerate}

Formally, the task is defined as learning or estimating two functions:
\[
\text{Sentiment}(a_i) \rightarrow \{-1, 0, +1\},
\]
\[
\text{Rationale}(a_i) \rightarrow \mathcal{P}(S_i),
\]
where $S_i$ is the set of sentences in article $a_i$, and $\mathcal{P}(S_i)$ denotes the power set over sentences (i.e., the subset selected as explanatory evidence).

%% file: dataset.tex
\section{Data Collection}
\label{sec:data_collection}
Our primary data source is the \textit{Media Cloud} platform. We have downloaded a list of URLs from there with a query
 that includes keywords `elephant', `jumbos', `hathi', `giants', and `tuskers'; timeline from `1 Jan 2022' to `30 Sept 2025'; from `The Outlet'\footnote{Name given by the authors in the interest of anonymity.}, which is among the most read English-language newspapers in India. From this query, we have collected 7,643 URLs along with metadata that includes \texttt{indexed\_date}, \texttt{language}, \texttt{media\_url}, \texttt{publish\_date}, \texttt{title}.

\subsection{Scraping Methodology and Technical Implementation}
The extraction was carried out using a custom Python script relying on the \texttt{newspaper} library for efficient article parsing and \texttt{BeautifulSoup} for custom element retrieval.\\
\textbf{Concurrent Processing:} A \texttt{ThreadPoolExecutor} (configured for a maximum of 10 concurrent workers) was utilized to parallelize URL processing, significantly reducing the total execution time and facilitating robust resumption from interruption.\\
\textbf{Core Article Extraction:} The \texttt{newspaper} library served as the primary tool to retrieve the \texttt{title} (headline) and the main \texttt{body\_text}.\\
\textbf{Subheadline Extraction:} We implemented a rule-based approach using \texttt{BeautifulSoup} to extract subheadline. A pre-defined dictionary (\texttt{site\_rules}) mapped the domain to unique HTML tags and class attributes (e.g., \texttt{h2}) identified as containing the subheadline.\\
\textbf{Text Construction and Cleaning:} The final text corpus was constructed by concatenating the \texttt{Headline}, \texttt{Subheadline}, and \texttt{Body Text}. A critical cleaning step was included to de-duplicate the subheadline from the article body if it was redundantly included by the core parser, ensuring accurate representation of the text structure.



\subsection{Dataset Refinement}
The next challenge was to identify which articles were actually relevant to our study, since the initial keyword search returned non-relevant results as well. A final filtering step was performed to isolate articles specifically discussing human-elephant dynamics (conflict, interaction, or conservation).\\
\textbf{Classification Methodology:} Gemini, a large language model (LLM), was employed as an automated classifier. The model was prompted to classify each article into two categories: Relevant or Non-Relevant. This approach provided superior contextual accuracy compared to simple keyword exclusion lists. The representative prompt we have used is:\\

\texttt{Analyze the following news article regarding possible human–elephant conflict. \\
Task 1: Determine if the article describes human–elephant conflict.
Respond with either "Relevant" or "Not Relevant".\\
Task 2: Extract up to five location names (state, district, village).
Respond exactly in the format:\\
Relevance: <Relevant/Not Relevant>
Location: <semicolon-separated list or `Location not specified'> }\\

This meticulously crafted prompt underwent automated filtering, and the final relevance dataset was refined to 1,968 articles, comprising 28,989 sentences.

\subsection{Data Annotation}
\label{sec:data_annotation}

To establish a human-evaluated benchmark for sentiment and rationale quality, we conducted a manual annotation study involving two linguistic experts with prior training in discourse analysis and media framing. From the full corpus of news articles collected over three years and nine months, we selected a stratified random sample comprising more than $5\%$ of the dataset. This yielded a set of $101$ articles, sampled uniformly across the entire time range to avoid temporal bias and to ensure that variation in reporting patterns was adequately represented.

For each article, annotators were first asked to assign an overall sentiment label indicating whether the article portrays elephants in a positive, neutral, or negative manner. Beyond categorical sentiment, annotators also provided an \textit{intensity score} on a 1--10 scale, allowing a finer-grained assessment of how strongly the article conveyed its sentiment. This dual-level annotation framework enables both coarse and nuanced interpretations of narrative framing.

To support interpretability, annotators were required to provide explicit justifications for their assigned sentiment. These justifications consisted of two components: a list of lexical items present in the article that match entries from our Negative Elephant Portrayal Lexicon (NEPL), and a set of sentences that most directly contributed to the perceived sentiment. The NEPL-based lexical extraction highlights domain-specific cues associated with aggressive or sensational portrayals of elephants, while sentence-level rationales allow for transparent linkage between sentiment judgments and textual evidence. In addition, annotators were instructed to indicate whether the article mentioned any human or elephant deaths; if so, they recorded the number of deaths referenced in the narrative. 

\subsection{NEgative Portrayal Lexicon (NEPL) Construction}
\label{sec:nepl_construction}
To capture domain-specific linguistic patterns associated with negative framing, we construct a \textit{Negative Elephant Portrayal Lexicon (NEPL)}. The purpose of NEPL is not to assign sentiment scores, but to provide an interpretable set of lexical cues that support both human annotation and rule-based analysis.

The lexicon was developed in a corpus-driven and annotation-informed manner. First, during manual inspection of articles and the annotation process, we identified recurring words and phrases used in conflict-related reporting, including aggression verbs, fear-inducing descriptors, and fatality indicators. These terms were curated rather than automatically extracted, based on their repeated usage in contexts portraying elephants as dangerous, intrusive, or destructive. Two linguistic experts then consolidated the candidates into a structured lexicon by grouping semantically similar expressions and removing ambiguous or context-dependent items. Particular emphasis was placed on multi-word expressions such as ``rogue elephant'', ``trampled to death'', and ``rampaging herd'', as these phrases contribute more strongly to sensational framing than isolated words.

NEPL was iteratively refined during the annotation study: annotators explicitly used the lexicon to highlight supporting words and phrases when assigning sentiment labels, thereby validating coverage and surfacing overlooked terms. The resulting lexicon is organized into six categories—aggression and violence, intrusion and invasion, destruction and damage, fear and panic, metaphorical or anthropomorphic labels, and conflict and hostility (Table~\ref{tab:nepl}).

Overall, NEPL serves as a lightweight, interpretable resource that bridges qualitative annotation and quantitative analysis, and is further used in our VADER+RegEx hybrid pipeline to resolve domain-specific sentiment ambiguity. We note that the lexicon is not intended to be exhaustive, but rather to capture high-frequency and high-impact framing cues observed in the dataset.

\begin{table*}

\centering
\setlength{\tabcolsep}{1mm}{
\begin{tabular}{p{3.2cm} p{12.5cm}}
\toprule
\textbf{Category} & \textbf{Representative Terms and Expressions} \\
\midrule

\textbf{Aggression and Violence} &
aggressive, aggressor, enraged, angry, ferocious, furious, violent, rogue, rampage, rampaging, raging, charged, charging, trampled, trampling, crushed, stomped, mauled, gored, struck, deadly, killer, murderous, bloodthirsty, stampede, plunder, raid, ran amok, hurl, marauded \\

\textbf{Intrusion and Invasion} &
invade, invaded, invading, invasion, intrude, intruded, intruding, encroached, trespassed, trespassing, stormed, broke into, entered forcefully, barged, overran, crop-raiding \\

\textbf{Destruction and Damage} &
destroyed, destruction, damaged, demolished, wrecked, ruined, flattened, uprooted, devastated, ravaged, smashed, shattered, havoc, chaos, turmoil, mayhem \\

\textbf{Fear and Panic} &
panic, terror, terrified, horror, nightmare, frightened, alarming, scary, fearsome, menace, threatening, horrifying, dreadful, scary incident, tragic encounter, dangerous, danger, flee, shock, encounter, run away, escape, narrow escape, on edge \\

\textbf{Metaphorical / Anthropomorphic Labels} &
menace, monster, beast, brute, villain, killer beast, rogue tusker, rogue elephant, wild beast, terror tusker, killer jumbo, menacing giant, lone elephant, lone tusker, pachyderm, wild tusker, loner \\

\textbf{Conflict and Hostility} &
clash, standoff, confrontation, battled, retaliated, counterattacked, hunted down, human-elephant war, rampaging herd, man-elephant conflict, tension, chased \\


\bottomrule
\end{tabular}

\caption{Representative terms from the Negative Elephant Portrayal Lexicon (NEPL). These expressions were compiled with expert input and frequently appear in media narratives portraying elephants as dangerous, violent, or threatening.}
\label{tab:nepl}
}
\end{table*}



\subsection{Dataset Statistics}
\label{sec:dataset_statistics}

Our final dataset comprises news articles collected from the major Indian English-language media outlet {`The Outlet'} and spans more than 3 years (January 2022 to September 2025)\footnote{We also attempted to obtain the dataset from other news outlets. However, the number of articles from those outlets that we could access is very low, which is why we limit ourselves to `The Outlet'.}. After scraping, text extraction, cleaning, and relevance filtering with an LLM-based classifier, the resulting corpus contains 1,968 full-length articles comprising 28,986 sentences. These articles exhibit substantial variation in length, reflecting differences in editorial styles, reporting depth, and narrative structure across media sources. To characterize the dataset, we computed standard corpus-level statistics, including mean, median, and variance in token counts per article.

Table~\ref{tab:dataset_stats} presents a concise summary of the key dataset statistics.

\begin{table}[t]
\centering
\begin{tabular}{lc}
\toprule
\textbf{Statistic} & \textbf{Value} \\
\midrule
Total number of scraped articles & 1,968 \\
Total number of sentences & 28,986\\
Languages & English \\
Time period covered & Jan 2022 -- Sep 2025 \\
\midrule
Mean word count & 322.65 \\
Median word count & 293.5 \\
Standard deviation of word count & 199.97 \\
Minimum article length (in words) & 22 \\
Maximum article length (in words) & 2,410 \\
\midrule
Number of manually annotated &\\
test articles & 101 \\
\bottomrule
\end{tabular}
\caption{Summary statistics of the curated human--elephant conflict news dataset. Token counts are computed after text extraction and cleaning.}
\label{tab:dataset_stats}
\end{table}

%% file: methods.tex
\section{Method Overview}
\label{sec:method_overview}

Our method consists of two stages: sentiment prediction for each full news article and the extraction of rationale sentences that explain the predicted sentiment. To estimate sentiment, we apply multiple modeling strategies independently, allowing us to compare how different computational paradigms interpret sentiment in long-form reporting on human-elephant conflict. Large Language Models (LLMs), such as Gemini~\cite{gemini2023} and Qwen~\cite{qwen2024}, are used in zero-shot settings where the entire article is provided together with an instruction prompt requesting a sentiment label. These models, with their extensive context windows and reasoning abilities, provide a powerful mechanism for interpreting nuanced or implicit sentiment cues.

To complement LLM-based inference, we evaluate a long-document transformer model, Longformer~\cite{beltagy2020longformer}, which supports sequences up to $4,096$ tokens through a sparse attention mechanism. This enables full-article classification without truncation, capturing sentiment indicators that may appear anywhere in the document. As a standard transformer baseline, we additionally include a RoBERTa-based sentiment classifier~\cite{siebert2020roberta}. Although limited to shorter input lengths, RoBERTa provides a useful comparison to long-context and LLM-based predictions. It requires truncation to its maximum input window of 512 tokens, enabling us to examine how shorter-context models perceive sentiment compared to long-context architectures. Finally, we incorporate a hybrid rule-based sentiment analysis pipeline that combines VADER~\cite{hutto2014vader} with domain-specific RegEx heuristics. VADER is a lexicon- and rule-driven polarity analyzer that produces a continuous compound sentiment score in the range $[-1,+1]$. We first apply VADER to the full article text. Based on the VADER score distribution, we have observed three regions. We have chosen thresholds based on this distribution and classify sentiment as positive when the compound score exceeds $+0.20$, negative when it falls below $-0.20$, and ambiguous otherwise. For articles in this ambiguous range $(-0.20, +0.20)$, we apply a second-stage RegEx-based analysis inspired by pattern-driven linguistic methods~\cite{jurafsky2000speech}. This stage detects explicitly negative framing through handcrafted patterns derived from the NEPL, including aggressive verbs, fear-inducing descriptors, and fatal outcome indicators. Articles with more than two matched negative NEPL patterns are classified as negative; otherwise, they are labeled neutral. This two-stage design allows the pipeline to preserve VADER’s general sentiment sensitivity while resolving domain-specific ambiguity using explicit linguistic cues.

Each model operates independently: the output of one model does not influence the others, and no ensembling is performed. This allows us to systematically compare how different sentiment analysis families behave on the same corpus. In addition to sentiment labels, we extract sentence-level rationales using an instruction-following LLM. The model receives the full article text and is asked to identify the sentences that contribute most strongly to the predicted sentiment. This step, using NEPL, enables interpretability and provides a qualitative foundation for understanding linguistic patterns that contribute to negative portrayals.

%% file: experimental_setup.tex
\section{Experimental Setup}
\label{sec:experimental_setup}

Each of the articles is retained in its original form without separating headlines, subheadlines, or body text. Preprocessing is intentionally minimal, consisting only of the removal of boilerplate HTML, normalization of Unicode characters, and sentence segmentation using \texttt{spaCy} to support rationale extraction. No additional linguistic transformations, such as stopword removal or lemmatization, are applied, preserving the authentic writing style and narrative tone of news reporting. 

For sentiment prediction, we evaluate five approaches: Gemini, Qwen, Longformer, RoBERTa, and VADER + RegEx-based classifier. These models differ in their native output formats: Gemini and Qwen generate discrete sentiment labels through instruction-following prompts; Longformer outputs five sentiment classes ranging from very negative to very positive; RoBERTa produces softmax probability distributions over sentiment classes; VADER returns a continuous compound polarity score in the range $[-1,+1]$; and the RegEx-based classifier emits rule-based decisions based on domain-specific negative framing patterns. To enable consistent comparison across models, all outputs are deterministically mapped into a unified sentiment space $\{-1,0,+1\}$ corresponding to negative, neutral, and positive sentiment. For Gemini and Qwen, the predicted label is directly mapped. For Longformer, very negative and negative classes are mapped to $-1$, neutral to $0$, and positive and very positive to $+1$. For RoBERTa, the class with maximum softmax probability is selected and mapped accordingly. For VADER, we have observed three clusters in the scores. We have chosen thresholds based on this clustered pattern. The compound scores $\leq -0.20$ are mapped to $-1$, scores between $-0.20$ and $+0.20$ to $0$ (considered ambiguous), and scores $\geq +0.20$ to $+1$. Finally, the RegEx classifier based on NEPL is used for the ambiguous articles (with scores between $-0.20$ and $+0.20$ by VADER). It assigns $-1$ when at least three negative or fear-inducing patterns from NEPL are detected, $0$ otherwise (Figures~\ref{fig:Vader} and~\ref{fig:regex} in Appendix). The prompt design and other model parameters are summarized in Tables~\ref{tab:prompt} and \ref{tab:hyperparameter}.

\begin{table*}[t]
\centering
\caption{Combined Summary of Prompt Designs Used Across All Models}
\label{tab:prompt}
\resizebox{\textwidth}{!}{
\begin{tabular}{p{2.0cm} p{13.5cm}}
\toprule
\textbf{Model} & \textbf{Prompt Structure and Core Instructions} \\
\midrule

\textbf{Gemini (API)} &
Task: Classify sentiment toward elephants and extract 2--5 supporting sentences.  
Includes explicit criteria for Positive, Neutral, Negative framing.  
Emphasis on aggression terms (e.g., ``terrorised'', ``invaded''), empathy cues, and balance of perspectives.  
Output strictly enforced as JSON:  
\{sentiment: ..., supporting\_sentences: [...]\}.  
Deterministic, JSON-only prompting. \\

\textbf{Qwen2.5–7B-Instruct} &
Role: “Expert news analyst specializing in Human–Elephant Conflict (HEC).’’  
Defines sentiment categories using examples (e.g., ``menace’’ → negative, ``rescue’’ → positive).  
Takes entire article as input.  
Output required as JSON with keys: classification, confidence, reasoning.  
Reasoning must be one short sentence. \\

\textbf{Longformer (5-class)} &
No natural-language prompt; the model is a direct sequence classifier.  
Input: raw article text (truncated to 4096 tokens).  
Output: one of five discrete labels: very negative, negative, neutral, positive, very positive. \\

\textbf{RoBERTa (Chunk-based)} &
No natural-language prompt (classification head).  
Chunk-level processing: 450-word windows with overlap.  
Elephant-context windows extracted using regex triggers (e.g., “elephant|tusker|jumbo’’).  
Final sentiment label determined via clustering over chunk-level negative scores. \\

\textbf{VADER + Regex Hybrid} &
No LLM prompt.  
Rule layer checks for negative-framing expressions (e.g., “killer’’, “rampage’’, “terror’’),  
victim patterns (deaths/injuries),  
and negation cues (``no casualties'', ``no injuries'').  
Combined with VADER’s lexicon-based compound score through handcrafted decision rules. \\
\bottomrule
\end{tabular}}
\end{table*}

\begin{table*}[t]
\centering
\caption{Combined Summary of Hyperparameters and Inference Settings for All Models}
\label{tab:hyperparameter}
\resizebox{\textwidth}{!}{
\begin{tabular}{p{2.0cm} p{13.5cm}}
\toprule
\textbf{Model} & \textbf{Key Hyperparameters and Inference Configuration} \\
\midrule

\textbf{Gemini (API)} &
Temperature: 0.0--0.2 (deterministic); Sampling disabled; Max tokens: 256;  
Response format: JSON-only; Runtime: Cloud-hosted (no local GPU);  
Lexicon: implicit (no manual list);  
Task: sentiment + supporting sentences;  
Prompt: structured instructions focusing on elephant portrayal;  
Error handling: retry loops for rate limits. \\

\textbf{Qwen2.5--7B-Instruct} &
Quantization: 4-bit NF4; Dtype: bfloat16; Max new tokens: 128; Temperature: 0.1;  
Context length: up to 32k tokens (handles full articles);  
Device: NVIDIA T4 GPU (Colab);  
Framework: HuggingFace chat template;  
Output: JSON containing classification, confidence, reasoning;  
Lexicon: implicit, inferred from definitions. \\

\textbf{Longformer (5-class)} &
Model: Muddassar/longformer-base-sentiment-5-classes;  
Max sequence length: 4096 tokens; Truncation: enabled; Batch size: 32;  
Device: GPU; Memory: 8--12GB depending on batch size;  
Task: 5-class sentiment (very neg → very pos);  
Lexicon: internal learned sentiment representation;  
Output: one sentiment label per article. \\

\textbf{RoBERTa (Chunk-based)} &
Model: siebert/sentiment-roberta-large-english;  
Max lengths: 512 (article chunks), 256 (context), 128 (headline);  
Chunk size: 450 words with 50-word overlap; Min chunk size: 20 words;  
Elephant-context sentences: ±1 window via regex;  
Clustering: KMeans (k=3, n\_init=10, random\_state=42) for final labels;  
Framework: Transformers + PyTorch; Execution: GPU. \\

\textbf{VADER + Regex Hybrid} &
Model: VADER (lexicon \& rule-based); Score: compound $\in[-1,1]$;  
Regex layers: fear terms (e.g., attacked, rampage, killer),  
victim patterns (deaths/injuries),  
negation patterns (no casualties);  
Decision logic: combines compound score + rule overrides;  
Output: vader\_compound, fear\_count, victim\_flag, final label;  
Execution: CPU (fast). \\

\bottomrule
\end{tabular}}
\end{table*}

Rationale extraction is conducted using an LLM that receives the full article and is instructed to return only those sentences that justify the sentiment prediction. To ensure consistency, the model outputs verbatim sentences extracted from the article itself. Since no gold-standard sentiment or rationale annotations exist for this dataset, evaluation is primarily qualitative. We compare the distribution of sentiment labels across all models, measure pairwise agreement between model predictions. Rationale coherence is assessed through manual inspection to determine whether the selected sentences genuinely reflect sentiment cues. This multi-faceted evaluation approach provides insights into how different models interpret the sentiment of conflict-related news and how their interpretations align with or diverge from one another.

%% file: results_and_analysis.tex
\section{Results and Analysis}
\label{sec:results}

This section presents overall sentiment trends, lexical evidence from NEPL, temporal and geographic variation, and details of agreement between AI tools and annotators.

\subsection{Sentiment distribution through NEPL}
\begin{figure}[t]
    \centering
    \includegraphics[width=\columnwidth]{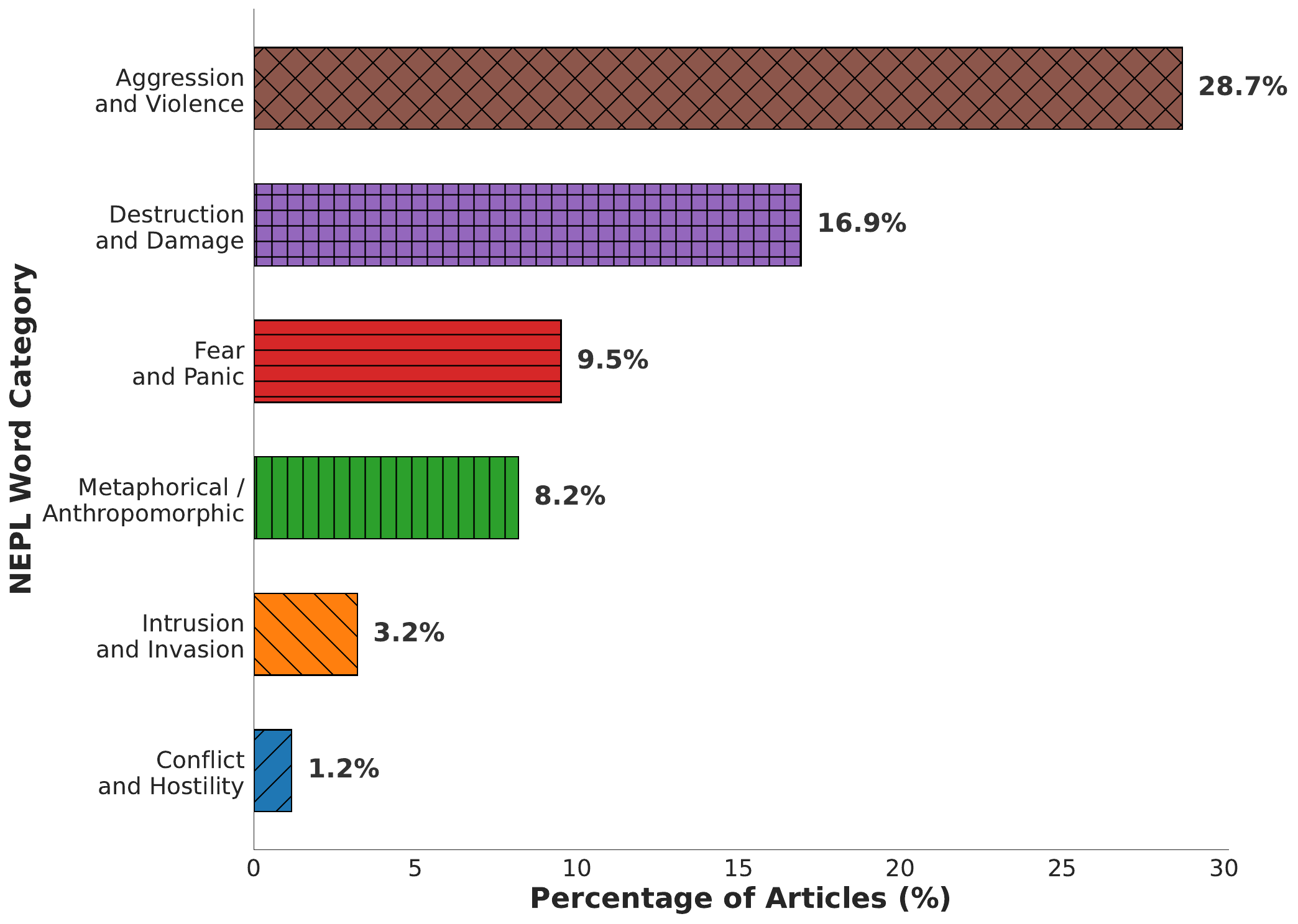}
    \caption{Presence of NEPL categories across the articles.}
    \label{fig:nepl_categories}
\end{figure}

Sentiment classification of the articles shows a dominant negative portrayal of elephants. Figure~\ref{fig:nepl_categories} summarizes the presence of NEPL categories in 1,971 articles. Note that each article may contain words from multiple categories. Surprisingly, a high proportion of articles (28.7\%) contain aggression and violence terms, followed by substantial presence of Destruction and damage (16.9\%). The word cloud is available in the appendix (Figure~\ref{fig:word_cloud}). 
These trends collectively demonstrate the prevalence of negative linguistic framing in HEC news coverage.

\subsection{Victim/Non-victim vs NEPL}
\begin{figure}[h]
    \centering
    \includegraphics[width=0.7\linewidth]{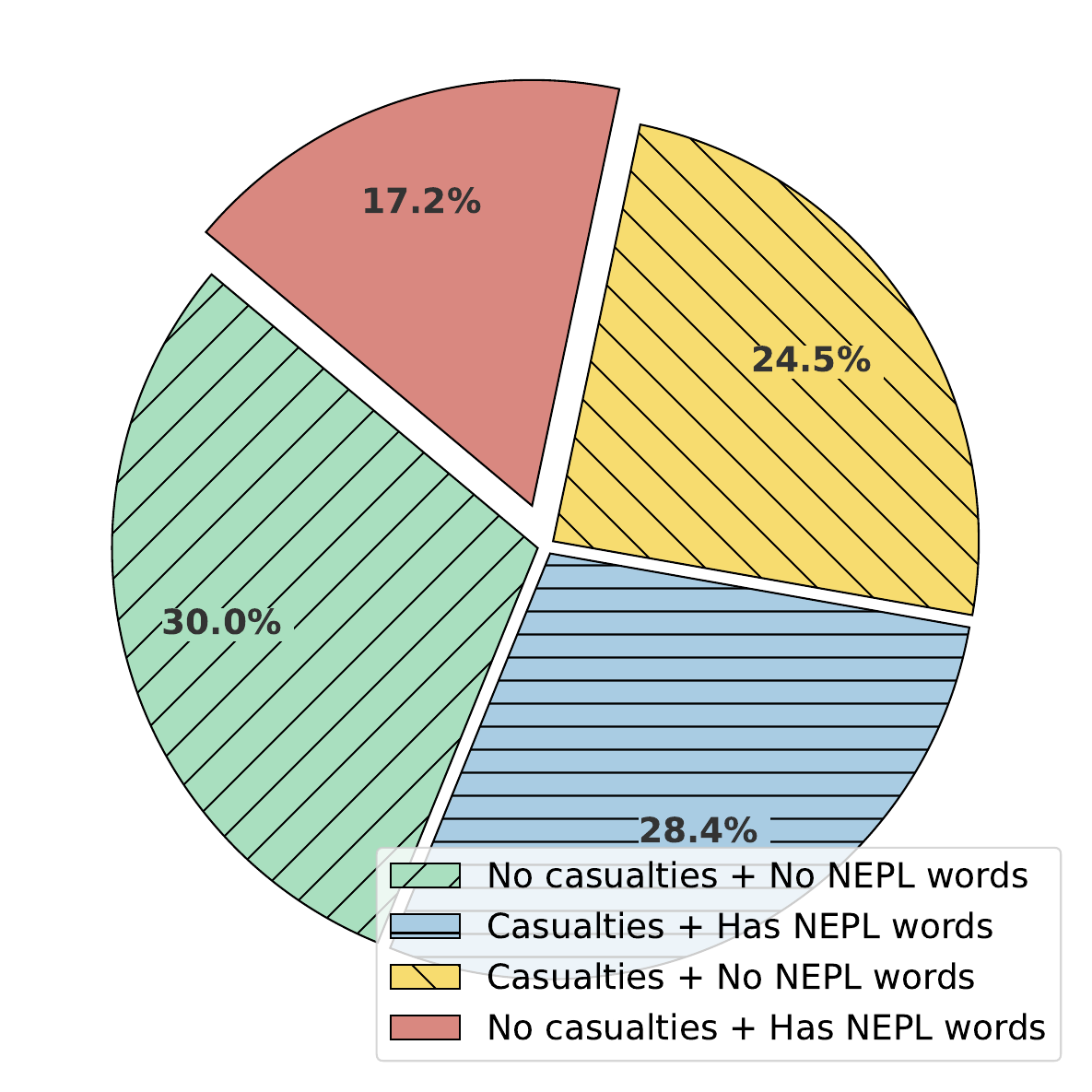}
    \caption{Mention of victim vs presence of NEPL}
    \label{fig:pie_chart}
\end{figure}

We question whether the negativity stems from the severity of the incidents. For that, we select a set of words (e.g., `killed', `killing', `fatal', `casualties', `trampled to death', `crushed to death', `lost lives', `trauma', `victim', `critical condition') whose occurrence reflects that the article does mention victims. Next, we check whether these articles have words from NEPL. The figure~\ref{fig:pie_chart} shows the four categories. More than 40\% of the articles contain NEPL words, and more than 17.2\% do not mention an incident of death or casualty but contain words from the NEPL categories. This indicates that, in some articles, harsh remarks could have been avoided while still conveying the severity of the incidents. 

\subsection{Cross-Model Convergence}
\begin{figure}[t]
    \centering
    \includegraphics[width=\columnwidth]{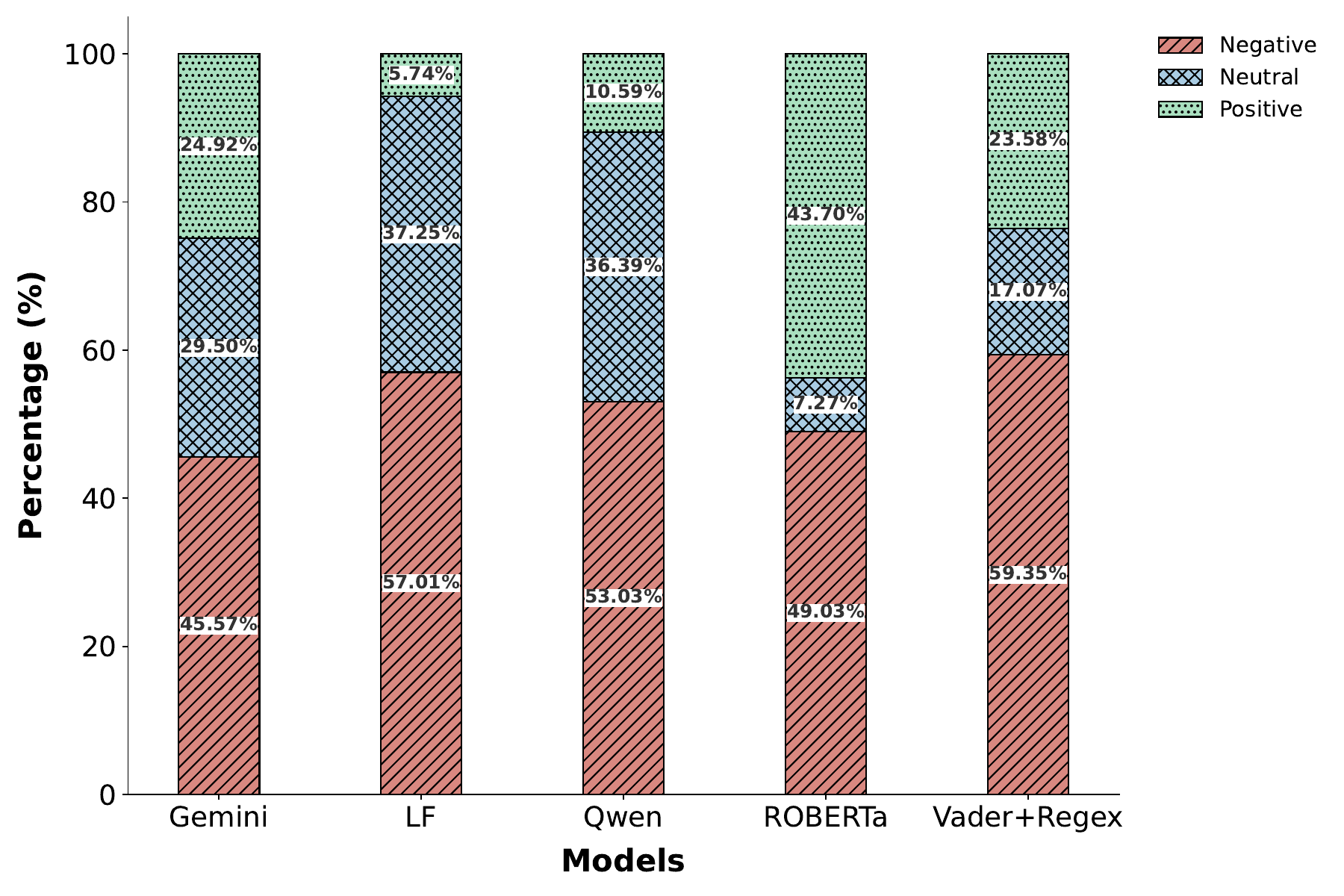}
    \caption{Comparison of proportion of labels across models. Despite methodological differences, all models classify the majority of articles as negative.}
    \label{fig:model_comp}
\end{figure}

A comparison of negativity rates across LLMs (Gemini, Qwen), long-context transformers (Longformer), standard transformers (RoBERTa), and lexicon-based models (VADER+Regex) shows strong cross-model agreement (Figure~\ref{fig:model_comp}). While LLMs and Longformer show slightly higher neutral proportions due to contextual sensitivity, all models consistently classify the majority of articles as negative. Lexicon-based models identify the highest negativity because they are sensitive to NEPL terms. This high convergence across modeling paradigms provides strong evidence that negative sentiment is inherent in the corpus rather than model-dependent.

\begin{figure}[t]
    \centering
    \includegraphics[width=\linewidth]{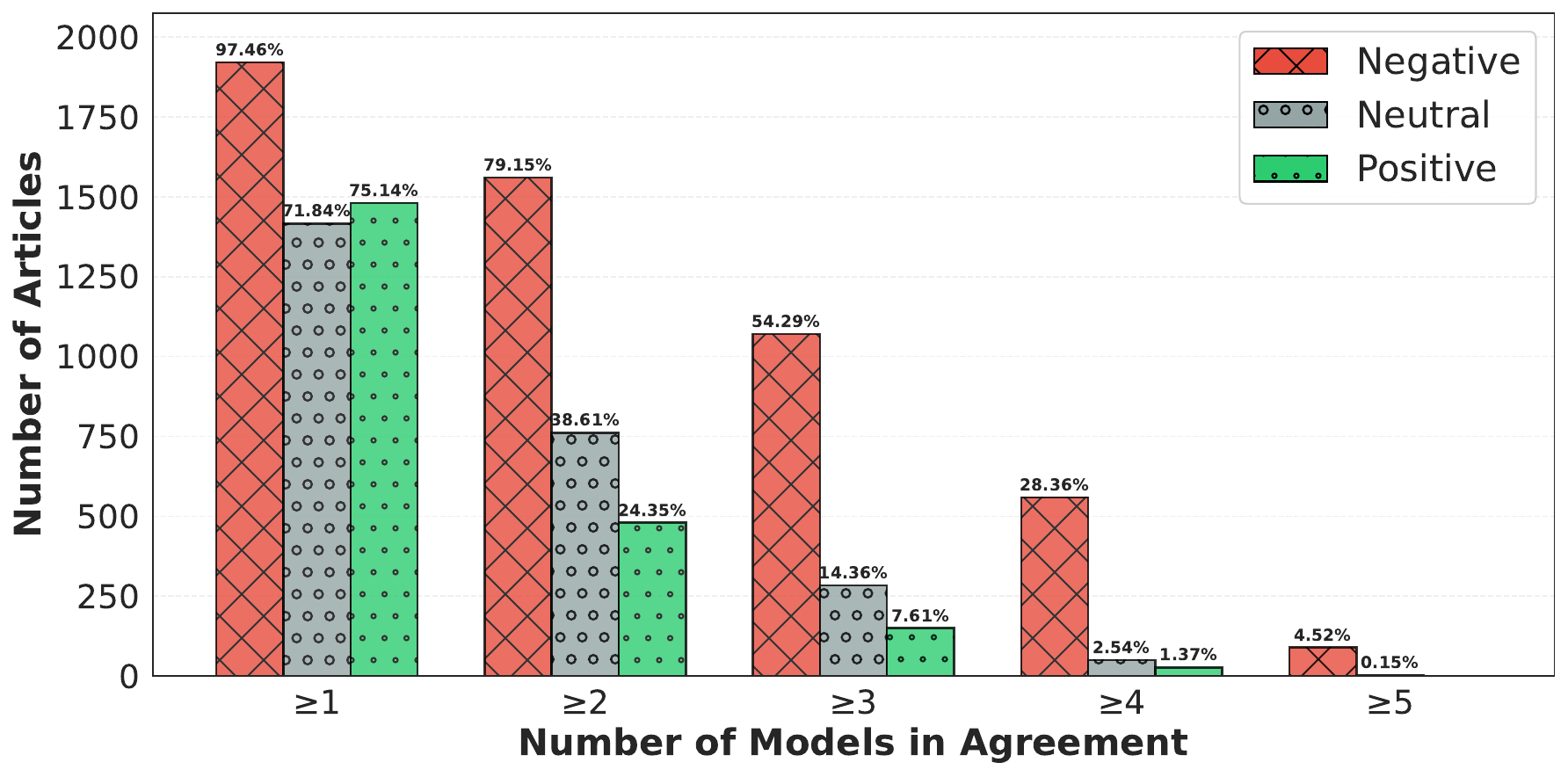}
    \caption{Comparison of negativity rates across LLM, transformer, and lexicon-based models. Despite methodological differences, all models classify the majority of articles as negative.}
    \label{fig:model_agreement_comp}
\end{figure}

We also analyse the agreement on the labeling done by the five models (Figure ~\ref{fig:model_agreement_comp}). Interestingly, around 54\% of the articles are labeled as negative by more than three out of five models. 

\subsection{Temporal Variation}
\begin{figure}[t]
    \centering
    \includegraphics[width=\columnwidth]{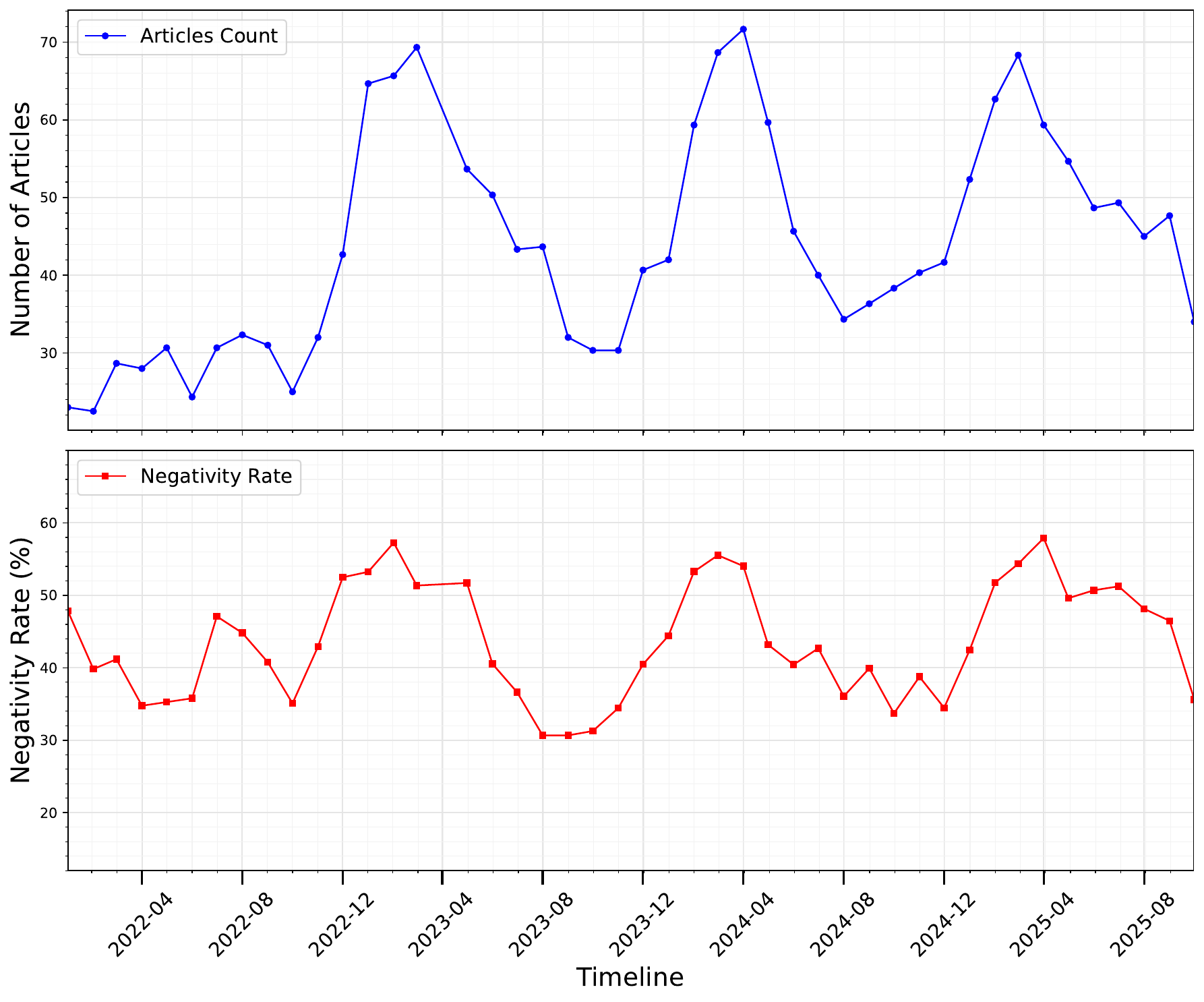}
    \caption{Monthly article count and the negativity rates across the full timeline  (Smoothed {window size 3}-month rolling mean) showing persistently high negative framing despite advisories from MoEF in Sep 2021 and then in Mar 2025.}
    \label{fig:monthly_neg}
\end{figure}

The monthly number of articles vs negative sentiment trends (Figure~\ref{fig:monthly_neg}) shows consistently significant negativity across the full timeline. To reduce random noise in the data while preserving underlying trends, we have constructed these plots after applying rolling mean smoothing with a window size of 3. There are noticeable spikes in both plots during January to April each year and a dip during the post-monsoon months (July to October) each year. This might be due to seasonality, for example, dry habitats, the timing of crop farming (e.g., paddy), and the flourishing of habitats in the post-monsoon.  

\subsection{AI vs. Human Expert Agreement}
\label{sec:llm_human_agreement}
To assess how closely AI models' predictions align with human judgment, we compare the models' sentiment labels with expert annotations. Two experts independently annotated 101 articles. They agreed on labels for 49 articles, resolved disagreements over the remaining articles, and provided final labels. For the 49 articles, the LLM (Gemini) achieves an agreement accuracy of 79.6\%, matching the expert sentiment labels in 39 of them. This relatively high alignment indicates that Gemini reliably captures dominant sentiment cues in human-elephant conflict reporting. A closer inspection reveals that agreement is strongest for negative sentiment. As shown in Table~\ref{tab:llm_vs_annotator_perf}, the Gemini achieves high precision (0.964) and recall (0.844) for negative articles, resulting in an F1-score of 0.900. This indicates that the model reliably captures fear-inducing and aggression-related framing commonly used in reporting HEC. Positive sentiment also exhibits strong recall (1.000), suggesting that Gemini successfully identifies empathetic or conservation-oriented narratives when explicitly present, although precision is lower due to occasional overgeneralization. Neutral sentiment remains the most challenging category, with comparatively lower precision (0.556) and recall (0.500), reflecting the inherent ambiguity of neutral framing and its reliance on subtle contextual balance rather than explicit lexical cues.

Disagreements between annotators and the LLM occur in 20.4\% of cases and are primarily concentrated around boundary conditions rather than polarity reversals. The most frequent mismatch involves articles annotated as negative being predicted as neutral by the LLM (40\% of disagreements), often when factual reporting of conflict events lacks explicit emotive language. Similarly, neutral articles are sometimes predicted as positive (40\% of differences). Importantly, extreme misclassifications are rare: direct flips from negative to positive account for only 10\% of differences, indicating that the LLM seldom contradicts expert judgment on overall sentiment direction.

\begin{table}[t]
\centering
\begin{tabular}{lcccc}
\toprule
\textbf{Sentiment} & \textbf{Precision} & \textbf{Recall} & \textbf{F1-Score} & \textbf{Support} \\
\textbf{Class} &  &  &  &  \\
\midrule
Negative & 0.964 & 0.844 & 0.900 & 32 \\
Neutral  & 0.556 & 0.500 & 0.526 & 10 \\
Positive & 0.583 & 1.000 & 0.737 & 7 \\
\bottomrule
\end{tabular}
\caption{Per-class performance of the Gemini compared against expert annotations on $49$ articles. Support denotes the number of articles annotated in each class.}
\label{tab:llm_vs_annotator_perf}
\end{table}

\textbf{Gemini vs Expert-annotated justification sentences.} Beyond sentiment label alignment, we evaluate whether the LLM identifies reasoning cues similar to those highlighted by human experts. Since understanding the justification for sentiment decisions is central to our task, we compare the sentences annotators marked as supporting evidence with the sentences the Gemini generated as rationales for its predictions. To quantify overlap between expert-selected and model-generated sentences, we employ standard lexical similarity metrics, BLEU and ROUGE-L, which capture n-gram precision and longest common subsequence overlap, respectively. 

The results indicate moderate agreement at the level of reasoning. As shown in Table~\ref{tab:llm_reasoning_agreement}, Gemini achieves higher overlap with Expert~2 than Expert~1, suggesting that while the model consistently captures salient negative framing cues, it may prioritize different but semantically related sentences when multiple supporting passages are present. These findings highlight that LLMs often converge with human experts on the core narrative drivers of sentiment, even when exact sentence selection differs.
\begin{table}[t]
\centering
\begin{tabular}{lcc}
\toprule
\textbf{Comparison} & \textbf{BLEU} & \textbf{ROUGE-L} \\
\midrule
Expert 1 vs. Gemini & 0.3849 & 0.5384 \\
Expert 2 vs. Gemini & 0.4784 & 0.5004 \\
\bottomrule
\end{tabular}
\caption{Lexical overlap between expert-annotated justification sentences and Gemini-generated reasoning.}
\label{tab:llm_reasoning_agreement}
\end{table}

A detailed result for per-class comparison of the AI model vs annotators on 101 articles is in Table~\ref{tab:models_vs_annotator_perf}.

\begin{table}[t]
\centering
\begin{tabular}{lcccc}
\toprule \textbf{Model} &
\textbf{Sentiment} & \textbf{Precision} & \textbf{Recall} & \textbf{F1-Score}\\
 &
\textbf{Class} &  &  & \\
\hline
\multirow{3}{4em}{Gemini} & 
Negative     &    0.68    &    0.74   &     0.71    \\
 &      Neutral   &      0.63    &    0.45    &    0.52 \\
  &    Positive     &    0.52    &    0.70    &    0.60 \\
\hline
\multirow{3}{4em}{Qwen} & 
  Negative     &   0.62     &  0.81    &   0.71 \\
 &  Neutral   &   0.56    &   0.47    &   0.51 \\
   &  Positive  &   0.69   &    0.45   &    0.55\\
\hline
\multirow{3}{4em}{RoBERTa} & 
Negative   &    0.18   &   0.19   &   0.18  \\
&     Neutral    &   0.33  &    0.08    &  0.13 \\
 &   Positive   &    0.02    &  0.05    &  0.03  \\ 
\hline
\multirow{3}{4em}{Regex} & 
 Negative  &0.61   &   0.88   &   0.72\\
& Neutral  & 0.57   &   0.32   &   0.41 \\
& Positive & 0.78    &  0.70     & 0.74\\
\hline
\multirow{3}{1em}{LF} & 
 Negative & 0.57 &  0.84 & 0.68 \\
& Neutral  & 0.53 & 0.45 & 0.49 \\
& Positive & 0.67 & 0.20 & 0.31 \\
\hline
\end{tabular}
\caption{Per-class performance of the models compared against expert annotations on the sentiment classification task for the complete $101$ articles. Support (number of articles labeled by experts in each class) is $43$, $38$, and $20$ for Negative, Neutral, and Positive, respectively.}
\label{tab:models_vs_annotator_perf}
\end{table}

%% file: discussion_combined.tex
\section{Discussion, Limitations, and Ethical Considerations}
\label{sec:discussion_combined}
India’s extensive print media ecosystem—one of the largest globally with over 110 million daily newspaper copies circulated~\cite{IR_survey2019}— that can play a central role in shaping public perceptions of wildlife and conservation. Competitive pressures and audience-driven incentives often lead outlets to prioritize dramatic or sensational narratives~\cite{de2012human}. Our analysis indicates that this broader media logic manifests strongly in reporting on human-elephant conflict (HEC), where negative portrayals far outweigh neutral or positive ones. Aggression-related and fear-inducing terminology appears in the vast majority of articles, reinforcing the perception of elephants as threatening actors rather than wildlife responding to ecological pressures such as habitat fragmentation and disrupted migratory routes. Such framings risk influencing public opinion, shaping mitigation policies, and undermining long-term coexistence efforts.

\textbf{Limitations of Current Advisories.} Despite advisories issued by India’s Ministry of Environment, Forest, and Climate Change (MoEF) in 2021 and 2025 discouraging sensational coverage, our results show little evidence of their impact. Often, news about advisories is not covered by the media. More effective change would require systematic engagement with media organizations, more straightforward editorial guidelines, journalist training, and mechanisms for accountability. Without such structural interventions, advisories alone are unlikely to shift entrenched reporting practices.

\textbf{Methodological Challenges.} This study, however, faces several methodological limitations. Sentiment cues in long articles may occur beyond the processing limits of standard transformer models, and although long-context models help address this, they impose higher computational costs and may still miss nuanced polarity. Linguistic signals contributing to negative portrayals vary widely: some are lexical (e.g., violent verbs), others contextual or event-driven (e.g., casualty descriptions). Lexicon- and rule-based systems capture only limited aspects of this spectrum, while large language models may introduce biases from their pretraining data. Moreover, evaluation is inherently challenging due to the absence of large-scale ground truth; expert annotations revealed moderate agreement, underscoring the subjective nature of sentiment and framing in journalistic narratives.

\textbf{Ethical Standards in Research:} We also adhered to ethical standards throughout the study. Articles were collected using unbiased search queries on an open-source platform, without manual filtering by author or region. Sentiment analysis was conducted solely on article text, excluding metadata such as journalist names or publication identifiers, ensuring that no individual or organization was targeted. Human annotators were compensated in accordance with institutional policy, and all annotations were treated confidentially. Our goal is not to critique specific journalists or outlets but to illuminate systemic linguistic patterns that may influence public understanding of HEC and conservation discourse.

Overall, the findings highlight the need for more balanced, context-aware reporting practices and emphasize the importance of methodological care and ethical sensitivity in computational media analysis.

%% file: conclusion_and_future_work.tex
\section{Conclusion and Future Directions}
This study provides a large-scale, data-driven examination of how Indian news media portray human-elephant conflict (HEC), revealing a consistent dominance of negative and fear-amplifying narratives across outlets. Through a combination of long-context transformers, large language models, rule-based systems, and a domain-specific Negative Elephant Portrayal Lexicon (NEPL), we demonstrate that sensational framing, rather than ecological context, often shapes public perception of elephants. These findings highlight the need for greater accountability and balance in media reporting, particularly in conservation-sensitive domains where narratives can influence policy and community attitudes. Moving forward, we aim to expand this work by incorporating multilingual reporting across Indian languages, integrating temporally aware models to analyze shifts in framing, and collaborating with ecologists and journalism experts to design proactive media guidelines. By releasing our dataset, lexicon, and tools, we hope to catalyze interdisciplinary efforts that advance responsible communication, support evidence-based conservation strategies, and ultimately foster more informed and empathetic human--elephant coexistence. A more comprehensive study could compare other regional media outlets, such as Hindi. Based on the comments from wildlife enthusiasts, we expect greater negativity and the use of sensational language in the regional media.

%% file: appendix.tex
\section{Appendix}

\subsection{Model Hyperparameters and Prompt Design}
\label{sec:hyperparameters}
We employed five different classes of models to compute sentiment scores and extract reasoning sentences across the human–elephant conflict news dataset. To ensure reproducibility, we report all implementation settings, decoding configurations, and prompt templates used for inference. For API-based models such as Gemini, deterministic decoding was enforced using low-temperature settings and disabled sampling, while local transformer models were executed on GPU-enabled environments with explicit control over quantization and context lengths. Because long-form news articles exhibit substantial variation in length, models such as Longformer and Qwen were selected for their extended context capabilities.

Each model family uses a distinct inference mechanism: (i) Gemini relies on structured JSON-only prompting with controlled generation; (ii) Qwen performs deterministic chat-style inference on quantized 7B parameters; (iii) Longformer conducts sequence classification over 4096-token windows; (iv) RoBERTa computes chunk-wise sentiment followed by cluster-based aggregation; and (v) the VADER+Regex hybrid combines lexicon-based scoring with handcrafted rule patterns to capture negativity markers specific to conflict reporting. All prompts and hyperparameters are provided in the table~\ref{tab:hyperparameter} and \ref{tab:prompt} for transparency and replication. Prompt used for Gemini is as follows:

\texttt{\small You are analyzing news articles about human–elephant conflict.\\
Article Title: {title}\\
Article Text: {excerpt}\\
Task:\\
Determine how elephants are portrayed.\\
Use the following criteria:\\
- NEGATIVE: Elephants portrayed using hostile language (e.g., "attacked", "invaded", "menace").\\
- NEUTRAL: Balanced reporting without emotionally charged terms.\\
- POSITIVE: Reporting that emphasizes ecological context, habitat loss, or empathy.}

Prompt used for Qwen is as follows:

\texttt{ \small You are an expert news analyst specializing in Human-Elephant Conflict (HEC).\\
Analyze the following news article and classify the portrayal of the elephant.\\
CLASSIFICATION DEFINITIONS:\\
- Negative:The article portrays elephants as a threat, 'killer', 'menace', or focuses on conflict/damage/fear.\\
-Positive: The article portrays elephants with sympathy, reverence (Ganesh), focuses on rescue/conservation, or is a lighthearted story.\\
- Neutral: Purely factual reporting (e.g., census numbers, policy updates) without emotional language.}

\section{VADER-Regex output}
\begin{figure}
    \centering
    \includegraphics[width=\linewidth]{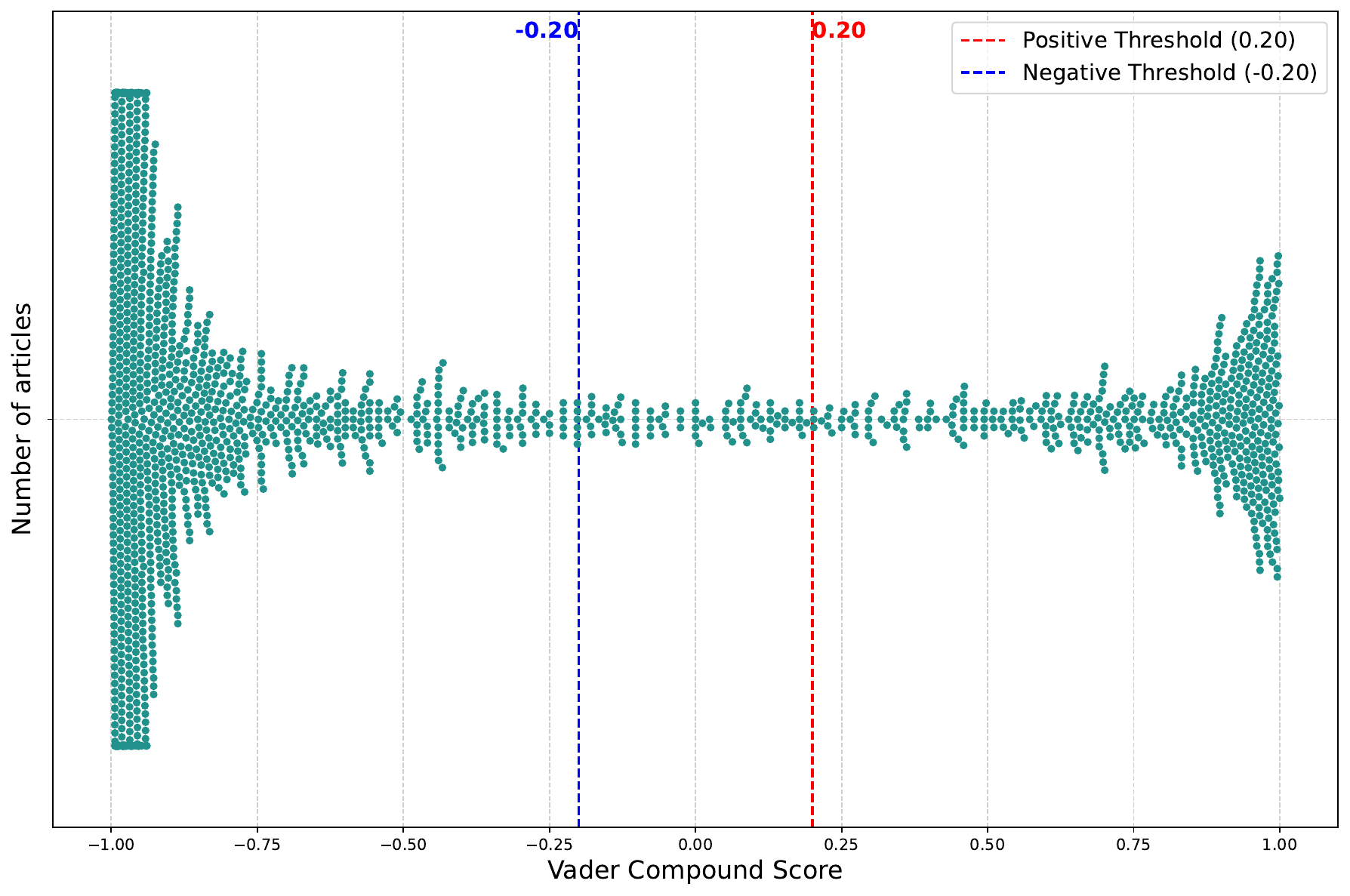}
    \caption{Vader Compound Scores vs number of articles}
    \label{fig:Vader}
\end{figure}

\begin{figure}
    \centering
    \includegraphics[width=\linewidth]{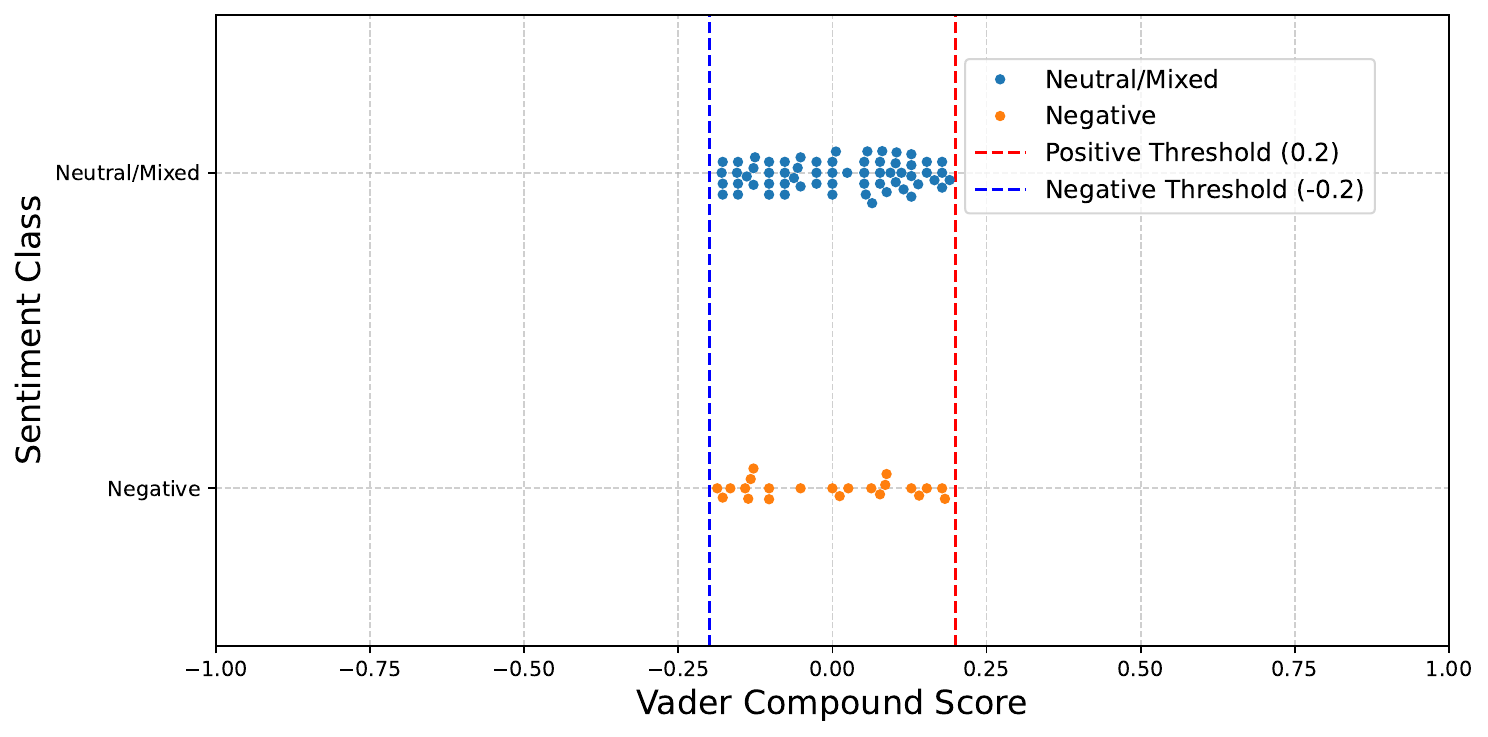}
    \caption{Output of Regex on articles having Vader Compound Scores between the negative and positive thresholds}
    \label{fig:regex}
\end{figure}

For VADER, we have observed three clusters in the scores. We have chosen thresholds based on this clustered pattern. The compound scores $\leq -0.20$ are mapped to $-1$, scores between $-0.20$ and $+0.20$ to $0$ (considered ambiguous), and scores $\geq +0.20$ to $+1$. Finally, the RegEx classifier based on NEPL is used for the ambiguous articles (with scores between $-0.20$ and $+0.20$ by VADER). It assigns $-1$ when at least three negative or fear-inducing patterns from NEPL are detected, $0$ otherwise (Figure~\ref{fig:Vader} and~\ref{fig:regex}).


\section{Sentiment distribution through NEPL}
\begin{figure}
    \centering
    \includegraphics[width=\linewidth]{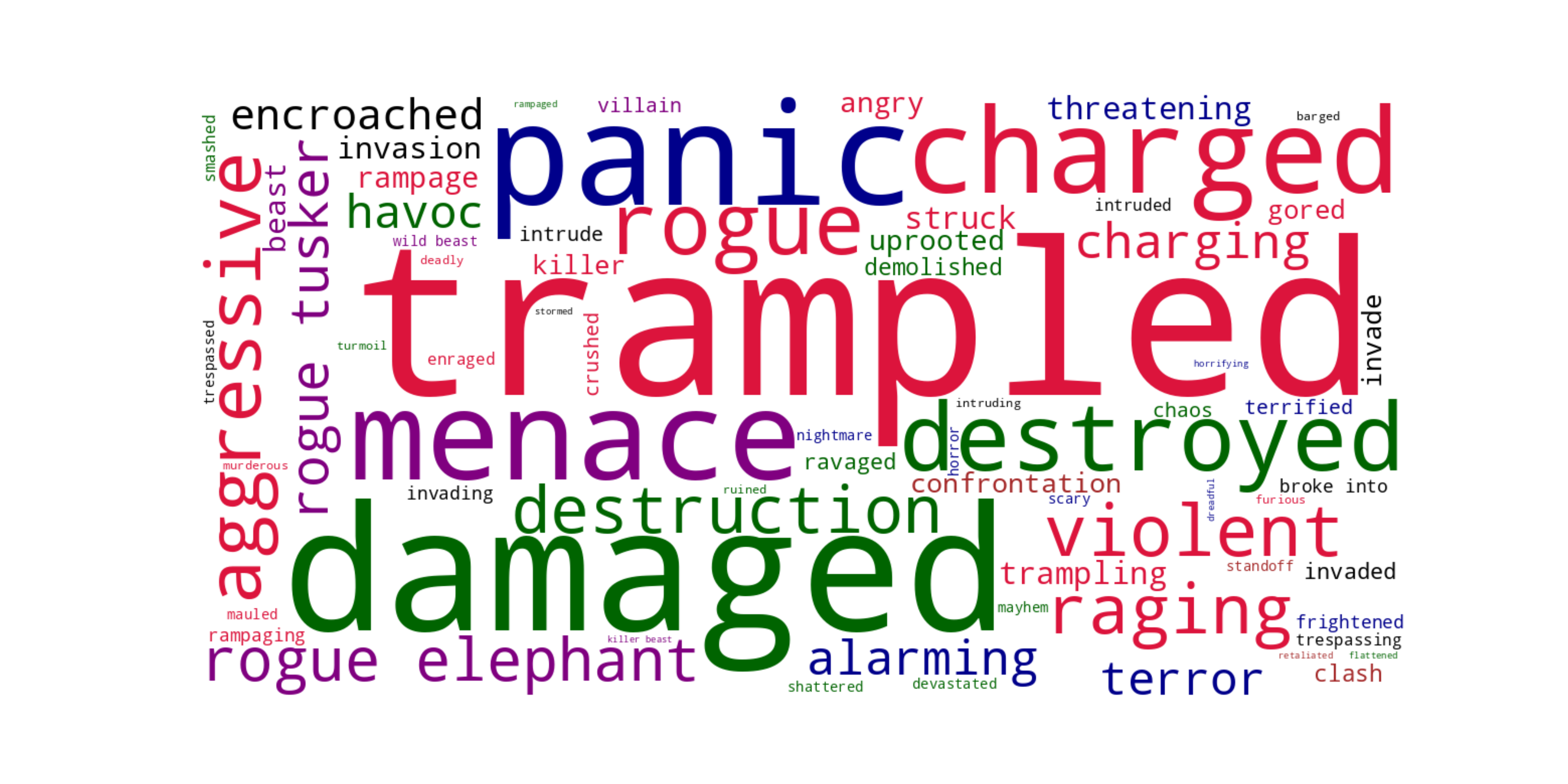}
    \caption{Word cloud to show sentiment distribution through NEPL}
    \label{fig:word_cloud}
\end{figure}

Sentiment classification of the articles shows a dominant negative portrayal of elephants. Figure~\ref{fig:nepl_categories} summarizes the presence of NEPL categories in 1,971 articles. Note that each article may contain words from multiple categories. Surprisingly, a high proportion of articles (28.7\%) contain aggression and violence terms, followed by substantial presence of Destruction and damage (16.9\%). These trends collectively demonstrate the prevalence of negative linguistic framing in HEC news coverage. The word cloud for most frequent words is available as figure~\ref{fig:word_cloud}. 

